\documentclass[conference]{IEEEtran}
\IEEEoverridecommandlockouts

\usepackage{cite}
\usepackage{amsmath,amssymb,amsfonts}
\usepackage{algorithmic}
\usepackage{graphicx}
\usepackage{textcomp}
\usepackage{xcolor}
\usepackage{enumitem}
\newlist{boldenum}{enumerate}{1}
\setlist[boldenum]{label=\textbf{(\arabic*)}}
\usepackage{setspace}
\usepackage{booktabs}
\usepackage{pdflscape}
\usepackage{stfloats}
\usepackage{multirow}
\usepackage{makecell}
\usepackage{pifont}
\usepackage{xr}
\usepackage{colortbl}
\usepackage[colorlinks=true, citecolor=blue, linkcolor=blue]{hyperref}
\usepackage{zref-xr}
\zxrsetup{toltxlabel}
\usepackage{tabularx} 
\usepackage{enumitem} 
\usepackage{booktabs} 
\usepackage{adjustbox}
\usepackage{caption}

\begin{document}

\title{HadaSmileNet: Hadamard fusion of handcrafted and deep-learning features for enhancing facial emotion recognition of genuine smiles
}

\author{\IEEEauthorblockN{Mohammad Junayed Hasan}
\IEEEauthorblockA{\textit{CS Department} \\ 
\textit{Johns Hopkins University}\\
Baltimore, MD, USA \\
mhasan21@jhu.edu}
\and
\IEEEauthorblockN{Nabeel Mohammed}
\IEEEauthorblockA{\textit{Apurba NSU R\&D Lab} \\
\textit{North South University}\\
Dhaka, Bangladesh \\
nabeel.mohammed@northsouth.edu}
\and
\IEEEauthorblockN{Shafin Rahman}
\IEEEauthorblockA{\textit{Apurba NSU R\&D Lab} \\
\textit{North South University}\\
Dhaka, Bangladesh \\
shafin.rahman@northsouth.edu}
\and
\IEEEauthorblockN{Philipp Koehn}
\IEEEauthorblockA{\textit{CS Department} \\
\textit{Johns Hopkins University}\\
Baltimore, MD, USA \\
phi@jhu.edu}
}

\maketitle

\begin{abstract}
The distinction between genuine and posed emotions represents a fundamental pattern recognition challenge with significant implications for data mining applications in social sciences, healthcare, and human-computer interaction. While recent multi-task learning frameworks have shown promise in combining deep learning architectures with handcrafted D-Marker features for smile facial emotion recognition, these approaches exhibit computational inefficiencies due to auxiliary task supervision and complex loss balancing requirements. This paper introduces HadaSmileNet, a novel feature fusion framework that directly integrates transformer-based representations with physiologically-grounded D-Markers through parameter-free multiplicative interactions. Through systematic evaluation of 15 fusion strategies, we demonstrate that Hadamard multiplicative fusion achieves optimal performance by enabling direct feature interactions while maintaining computational efficiency. The proposed approach establishes new state-of-the-art results for deep learning methods across four benchmark datasets: UvA-NEMO (88.7\%, +0.8\%), MMI (99.7\%), SPOS (98.5\%, +0.7\%), and BBC (100\%, +5.0\%). Comprehensive computational analysis reveals 26\% parameter reduction and simplified training compared to multi-task alternatives, while feature visualization demonstrates enhanced discriminative power through direct domain knowledge integration. The framework's efficiency and effectiveness make it particularly suitable for practical deployment in multimedia data mining applications that require real-time affective computing capabilities.
\end{abstract}


\section{Introduction}\label{sec:intro}

Facial expressions, as a universal form of non-verbal communication, play an essential role in social interactions, emotional well-being, and human-computer interfaces \cite{dong2022intentional,barrett2016works}. Within the spectrum of facial expressions, accurately distinguishing between genuine (Duchenne) and posed (non-Duchenne) smiles is a critical challenge, as it involves subtle, often imperceptible differences in muscle activations \cite{ekman1990duchenne, wegrzyn2017mapping}. The capability of reliably identifying genuine smiles has significant implications for multimedia data mining applications across diverse domains including human-computer interaction, social robotics, psychological research, healthcare diagnostics, and marketing strategies \cite{sarma2021methods, BRUCE201595, oh2016let, lander2020recognizing, li2020deep}, underscoring its wide-ranging importance and driving a growing body of research in pattern recognition and related fields.

\begin{figure}
    \centering
    \includegraphics[width=1\linewidth]{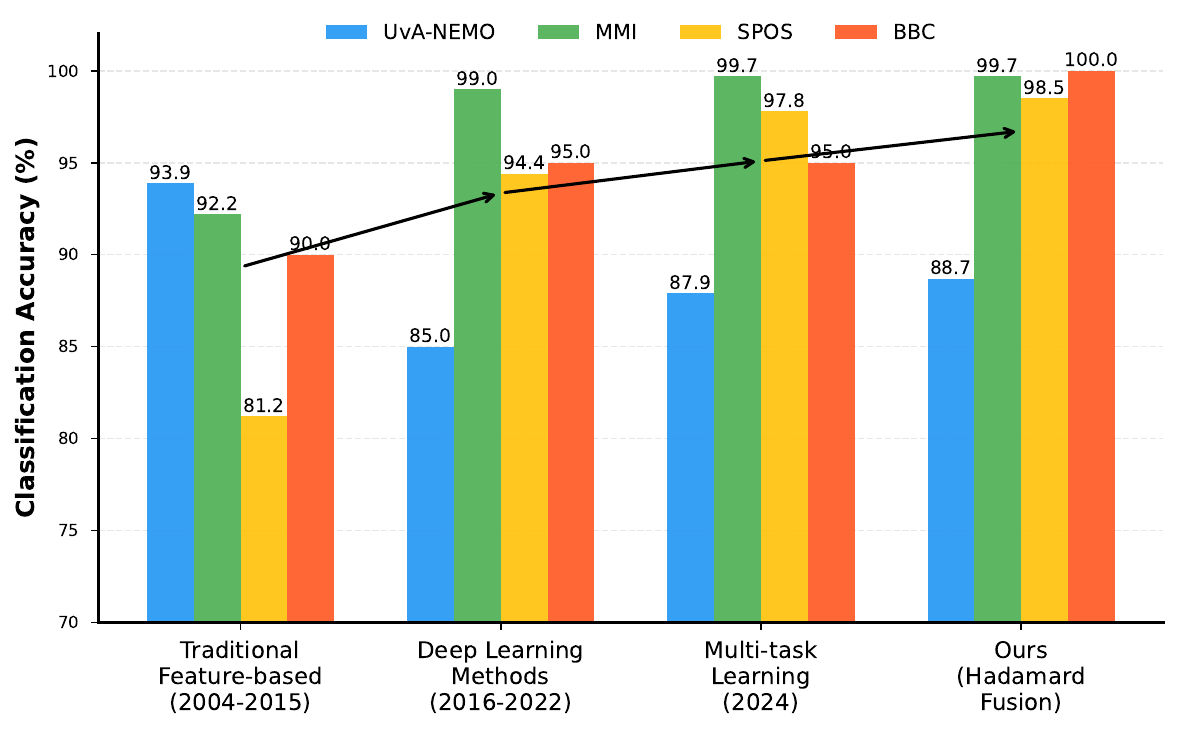}
    \caption{Evolution of smile facial emotion recognition methods: Our approach surpasses all end-to-end models across four benchmark datasets: UvA NEMO, MMI, SPOS, and BBC.}
    \label{fig:motivation}
\end{figure}

Figure \ref{fig:motivation} illustrates the comparative best performance of different existing methods, and places our proposed method in the literature. Early approaches for genuine smile recognition have often focused on hand-engineered, manually annotated features \cite{kawulok2021dynamics, hassen2021new, ratnawati2019features, cohn2004timing, dibeklioglu2010eyes, pfister2011differentiating, wu2014spontaneous, dibekliouglu2015recognition, wu2017spontaneous, mandal2017spontaneous}, most notably, the Duchenne Marker (D-Marker) features of the Facial Action Coding System (FACS) that highlight subtle facial muscular movements commonly associated with spontaneous or genuine emotions \cite{ekman1978facial, ekman1993facial}. Although effective, these hand-crafted methods are labor intensive, brittle to variations in data (e.g., illumination, pose), and often require significant domain expertise to ensure relevance and quality. With the emergence of deep learning, researchers have turned to architectures such as convolutional neural networks (CNNs), recurrent neural networks (including LSTMs), and more recently vision transformers (ViTs) \cite{mandal2017distinguishing, yang2020realsmilenet, faroque2022less, fan2022pstnet, fan2021point, dosovitskiy2020image}, thus achieving automatic feature extraction and surpassing traditional methods. Although these end-to-end learning strategies reduce manual efforts, they often disregard the rich, domain-relevant cues embedded in D-Markers, limiting their interpretability and potentially leaving valuable discriminative information untapped. To handle this limitation, a multi-task learning (MTL) framework, DeepMarkerNet \cite{HASAN2024148}, was recently proposed that formulates smile recognition as a multi-task problem, training one head to predict D-Markers and another to classify smiles, then discarding the former at test time. Although this strategy injects physiological knowledge, it introduces an additional classification head, inflates parameter count, and necessitates careful loss-weight tuning, factors that complicate training and reproducibility. The coupling of D-Marker prediction and smile classification may not fully exploit the complementary nature of these signals, as MTL frameworks often treat D-Marker prediction as a secondary supervisory cue rather than as a core part of the feature space. Furthermore, the indirect usage of D-Markers can hinder the model's capacity to learn robust joint representations, while inference-stage complexity remains similar to other deep learning models, offering limited additional interpretability or adaptability despite the added complexity in training.

In this paper, we propose \textbf{HadaSmileNet}—a novel framework that directly fuses D-Marker features with deep transformer-based representations for genuine smile recognition. Unlike MTL approaches that treat D-Markers as a learning signal to be predicted, we incorporate them at the feature level, allowing our model to leverage these domain-specific descriptors more explicitly. To achieve this, we systematically evaluate a comprehensive set of 15 simple and advanced feature fusion mechanisms, including various attention-based, bilinear, and multiplicative strategies. Among these, the Hadamard multiplicative fusion \cite{10965485} emerges as the most effective, offering a straightforward yet powerful means of blending domain knowledge with the richer, data-driven features learned by modern transformer architectures. By directly combining handcrafted and learned representations, our approach mitigates the training complexities, suboptimal weighting strategies, and under-realized relationships that characterize MTL frameworks. Moreover, similar to the MTL framework \cite{HASAN2024148}, we use feature fusion in the training phase only, and leave them out during inference, ensuring fair comparison and simplicity. We evaluate the framework on all available smile databases, and achieve state-of-the-art performance on all of them while reducing computational overhead, making it well-suited for practical multimedia data mining applications.

The key contributions of this study are as follows:
\begin{itemize}
    

    \item We introduce the first parameter-free feature fusion framework for genuine smile recognition that directly integrates physiologically-grounded D-Markers with transformer representations, eliminating the computational overhead and training complexity of multi-task learning approaches.
    \item Through systematic evaluation of 15 fusion strategies, we establish Hadamard multiplicative fusion as the optimal mechanism for domain knowledge integration, achieving state-of-the-art performance across four benchmark datasets with 26\% parameter reduction compared to existing methods.
    \item Comprehensive computational analysis demonstrates significant efficiency gains in training time, inference speed, and memory requirements, making the approach suitable for practical deployment in multimedia data mining applications requiring real-time processing capabilities.
\end{itemize}

\section{Methodology}\label{sec:method}

\subsection{Problem formulation}\label{sec:problem_formulation}

Let \(\{\mathbf{X}_i, y_i\}_{i=1}^{N}\) denote a collection of \(N\) video samples, where each sample \(\mathbf{X}_i = \{ \mathbf{x}_1, \mathbf{x}_2, \ldots, \mathbf{x}_{T_i}\}\) is a sequence of \(T_i\) facial frames. The label \(y_i \in \{0,1\}\) indicates whether the \(i\)-th video contains a posed (\(y_i=0\)) or a genuine (\(y_i=1\)) smile. Our primary objective is to learn a classification function \( f: \mathbf{X}_i \mapsto y_i \) that accurately discriminates between these two classes.

In addition to the raw video frames, we assume the availability of a handcrafted feature vector \(\mathbf{Z}_i \in \mathbb{R}^{M}\) for each video \(i\). This \(\mathbf{Z}_i\) encodes D-Marker features—highly discriminative handcrafted descriptors known to capture subtle facial muscle activations associated with genuine smiles. While previous approaches have either relied on these D-Markers as direct inputs or as auxiliary supervisory signals in a multi-task setting, our goal is to integrate these features more directly at the representation level.

Formally, let \(\mathcal{F}\) be a transformer-based feature extractor that processes the raw frames \(\mathbf{X}_i\) to produce a learned visual representation \(\mathbf{H}_i \in \mathbb{R}^{D}\):
\begin{equation}
    \mathbf{H}_i = \mathcal{F}(\mathbf{X}_i).
\end{equation}

We define a fusion operator \(\otimes: \mathbb{R}^{D} \times \mathbb{R}^{M} \mapsto \mathbb{R}^{Q}\) that combines the learned representation \(\mathbf{H}_i\) and the handcrafted D-Marker features \(\mathbf{Z}_i\), producing a fused feature vector \(\mathbf{F}_i \in \mathbb{R}^{Q}\):
\begin{equation}
    \mathbf{F}_i = \mathbf{H}_i \otimes \mathbf{Z}_i.
\end{equation}

Our classifier \( C: \mathbb{R}^{Q} \mapsto [0,1]\) then predicts the probability of the smile being genuine:
\begin{equation}
    \hat{y}_i = C(\mathbf{F}_i).
\end{equation}

The fusion operator \(\otimes\) can be implemented using various strategies, such as concatenation, attention-based weighting, bilinear pooling, or element-wise multiplicative integration. In this work, we systematically explore multiple such fusion techniques and identify those that most effectively combine the complementary information from \(\mathbf{H}_i\) and \(\mathbf{Z}_i\).

Training the model involves optimizing the parameters of \(\mathcal{F}\), \(\otimes\), and \(C\) to minimize a suitable loss function \(\mathcal{L}\), which in our case is the binary cross-entropy:
\begin{equation}
    \mathcal{L}(\hat{y}_i, y_i) = -\bigl[ y_i \log(\hat{y}_i) + (1-y_i)\log(1-\hat{y}_i) \bigr].
\end{equation}

By jointly learning \(\mathcal{F}\), \(\otimes\), and \(C\), the model can exploit both learned representations from a powerful transformer backbone and domain-specific D-Marker cues. This direct fusion paradigm eschews the complexities and constraints of multi-task frameworks, paving the way for more interpretable, efficient, and effective genuine smile classification models.
\begin{table}[!t]
\centering
\footnotesize
\caption{Mapping of the facial feature points to the indices of the AttentionMesh facial landmark extractor.}
\label{tab:landmarks_mapping}
\begin{tabularx}{\columnwidth}{XX}
            \toprule
            \textbf{Facial Point} & \textbf{Attention Mesh Index} \\ \midrule
            Right-eye outer corner & 33 \\ 
            Right-eye center & 159 \\
            Right-eye inner corner & 133 \\
            Left-eye outer corner & 362 \\
            Left-eye center & 386 \\
            Left-eye inner corner & 263 \\
            Right cheek & 50 \\
            Left cheek & 280 \\ 
            Nose tip & 1 \\ 
            Right lip-corner & 62 \\ 
            Left lip-corner & 308 \\
            \bottomrule
        \end{tabularx}
\end{table}

\subsection{Handcrafted feature extraction}\label{sec:handcrafted_features}

The handcrafted D-Marker features are derived following the physiologically-grounded approach established by Dibeklioğlu \textit{et al.} \cite{dibekliouglu2015recognition}, whose empirical validation demonstrated the discriminative power of these features for genuine smile recognition. The extraction process involves three key stages: facial landmark detection and selection, geometric preprocessing, and temporal D-Marker computation across three anatomically relevant facial regions.

\noindent\textbf{\textit{Landmark detection and selection.}} Raw video samples \(\mathbf{X}_i\) are processed through the off-the-shelf AttentionMesh model \cite{grishchenko2020attention}, an automated facial mesh prediction system developed by Google MediaPipe. This model was pre-trained on approximately 30,000 manually annotated facial images spanning diverse ethnicities and demographic groups, ensuring robust landmark detection across varied population groups. No additional fine-tuning is performed on AttentionMesh in our pipeline; only automatic inference is utilized to extract 478 3D facial landmark points through real-time tracking.

Following the protocol established by Dibeklioğlu \textit{et al.} \cite{dibekliouglu2015recognition}, we select a subset of $M=11$ key landmarks that correspond to facial action units AU6 (cheek raiser) and AU12 (lip corner puller), which are fundamental components of the Duchenne smile according to the Facial Action Coding System. The selection is algorithmic and based on anatomical relevance, requiring no manual annotation at inference time. These landmarks specifically target the contraction and movement of facial muscles orbicularis oculi and zygomaticus major, as well as subtle cheek and lip-corner motions critical for distinguishing genuine from posed smiles. Table~\ref{tab:landmarks_mapping} provides the mapping of the selected landmarks to their corresponding indices in the 478-point AttentionMesh topology.

To ensure consistent geometric interpretation, the selected landmarks of each frame are aligned to a reference coordinate system. Let $\mathbf{p}^{(x)}_{j} \in \mathbb{R}^3$ denote the 3D coordinates of the $j$-th selected landmark in frame $x$. We compute a plane using eye and nose reference points, and derive a normal vector to this plane. Using this normal, we estimate and remove head rotations (roll, yaw, pitch), followed by scale and translation adjustments. This normalization ensures that extracted features are robust to head pose variations and camera viewpoints.

\noindent\textbf{\textit{D-Marker computation.}} Once the landmarks are normalized, we derive three key dynamic measurements that collectively capture the structure of a smile over time:

\begin{enumerate}[leftmargin=1.2em]
    \item \textit{Lip Dynamics} ($D_{\text{lip}}$): Measures the relative distance and angular changes of lip corners from a stable reference, reflecting mouth opening and lip pulling (Eq. \ref{eq:5}).
    \item \textit{Eye Aperture} ($D_{\text{eye}}$): Quantifies eyelid opening or closing, capturing the hallmark crinkling around the eyes associated with genuine smiles (Eq. \ref{eq:6}).
    \item \textit{Cheek Elevation} ($D_{\text{cheek}}$): Tracks the vertical displacement of cheek regions, which lift prominently during genuine smiles (Eq. \ref{eq:7}).
\end{enumerate}

\begin{table}[t]
\caption{Facial feature definitions and total number of features extracted for each group. Features are derived from the empirically validated study by Dibeklioğlu \textit{et al.} \cite{dibekliouglu2015recognition}.}
\label{tab:features}
\centering
\footnotesize
\begin{tabular}{lcr}
\toprule
Feature & Definition & Features \\
\midrule
Duration & $\left[\frac{\eta(\mathcal{D}^+)}{\omega}, \frac{\eta(\mathcal{D}^-)}{\omega}, \frac{\eta(\mathcal{D})}{\omega}\right]$ & 3 \\[2ex]
Duration Ratio & $\left[\frac{\eta(\mathcal{D}^+)}{\eta(\mathcal{D})}, \frac{\eta(\mathcal{D}^-)}{\eta(\mathcal{D})}\right]$ & 2 \\[2ex]
Maximum Amplitude & $\max(\mathcal{D})$ & 1 \\[1ex]
Mean Amplitude & $\left[\frac{\sum\mathcal{D}}{\eta(\mathcal{D})}, \frac{\sum\mathcal{D}^+}{\eta(\mathcal{D}^+)}, \frac{\sum|\mathcal{D}^-|}{\eta(\mathcal{D}^-)}\right]$ & 3 \\[2ex]
STD of Amplitude & $\text{std}(\mathcal{D})$ & 1 \\[1ex]
Total Amplitude & $\left[\sum\mathcal{D}^+, \sum|\mathcal{D}^-|\right]$ & 2 \\[1ex]
Net Amplitude & $\sum\mathcal{D}^+ - \sum|\mathcal{D}^-|$ & 1 \\[2ex]
Amplitude Ratio & $\left[\frac{\sum\mathcal{D}^+}{\sum\mathcal{D}^++\sum|\mathcal{D}^-|}, \frac{\sum|\mathcal{D}^-|}{\sum\mathcal{D}^++\sum|\mathcal{D}^-|}\right]$ & 2 \\[2ex]
Maximum Speed & $\left[\max(\mathcal{V}^+), \max(|\mathcal{V}^-|)\right]$ & 2 \\[1ex]
Mean Speed & $\left[\frac{\sum\mathcal{V}^+}{\eta(\mathcal{V}^+)}, \frac{\sum|\mathcal{V}^-|}{\eta(\mathcal{V}^-)}\right]$ & 2 \\[2ex]
Max. Acceleration & $\left[\max(\mathcal{A}^+), \max(|\mathcal{A}^-|)\right]$ & 2 \\[1ex]
Mean Acceleration & $\left[\frac{\sum\mathcal{A}^+}{\eta(\mathcal{A}^+)}, \frac{\sum|\mathcal{A}^-|}{\eta(\mathcal{A}^-)}\right]$ & 2 \\[2ex]
Ampl. Duration Ratio & $\frac{(\sum\mathcal{D}^+-\sum|\mathcal{D}^-|)\omega}{\eta(\mathcal{D})}$ & 1 \\[2ex]
Ampl. Difference & $\frac{|\sum\mathcal{D}_L-\sum\mathcal{D}_R|}{\eta(\mathcal{D})}$ & 1 \\
\midrule
\multicolumn{2}{r}{\textbf{Total Number of Features:}} & \textbf{25} \\
\bottomrule
\end{tabular}
\end{table}

\begin{gather}
\label{eq:5}
D_{\text{lip}}(x) = \frac{\gamma(\frac{p_{10}^{1} + {p_{11} ^ 1}}{2}, p_{10}^x) + \gamma(\frac{p_{10}^{1} + {p_{11} ^ 1}}{2}, p_{11}^x)}{2\gamma(p_{10}^{1}, p_{11} ^ 1)} ,
\end{gather}
{\small{
\begin{gather}
\label{eq:6}
{D_{\text{eye}}(x) =
\frac{\Gamma(\frac{p_{1}^{x} + {p_{3} ^ x}}{2}, p_{2}^x)\gamma(\frac{p_{1}^{x} + {p_{3} ^ x}}{2}, p_{2}^x) + \Gamma(\frac{p_{4}^{x} + {p_{6} ^ x}}{2}, p_{5}^x)\gamma(\frac{p_{4}^{x} + {p_{6} ^ x}}{2}, p_{5}^x)}{2\gamma(p_{1}^{x}, p_{3} ^ x)}} ,
\end{gather}}}

\begin{gather}
\label{eq:7}
D_{\text{cheek}}(x) = \frac{\gamma(\frac{p_{7}^{1} + {p_{8} ^ 1}}{2}, p_{7}^x) + \gamma(\frac{p_{7}^{1} + {p_{8} ^ 1}}{2}, p_{8}^x)}{2\gamma(p_{7}^{1}, p_{8} ^ 1)} ,
\end{gather}

where, $\mathbf{p_{i}^{x}}$ represents the landmark at index $i$ in frame $x$, $\gamma()$ represents the Euclidean distance, and
$\Gamma(\mathbf{p_i}, \mathbf{p_j)}$ represents the relative vertical location, which equals -1 if $\mathbf{p_j}$ is located vertically below $\mathbf{p_i}$ on the face, and 1 otherwise.

For each frame $x$, these metrics $D_{\text{lip}}(x)$, $D_{\text{eye}}(x)$, and $D_{\text{cheek}}(x)$ are computed relative to baseline configurations observed in the initial frames. Differences in these values over time encode the temporal evolution of the smile. 

\noindent\textbf{\textit{Temporal Feature Aggregation.}} The three key phases of the smile, such as longest increasing segment (onset), stable apex interval, and longest decreasing segment (offset), are identified from the D-Marker metrics using the approach proposed in \cite{schmidt2003signal}. Each phase is again divided into increasing and decreasing segments for detailed analyses. From these segments, we derive a comprehensive set of temporal descriptors: duration-related measures, amplitude magnitudes, velocity and acceleration cues, and various ratios that capture the pattern of smile in the videos. As shown in Table \ref{tab:features}, a total of 25 features are calculated for each of the three phases, giving 75 features for each of the three facial regions (eyes, lips and cheeks). By concatenating all these features we form a final $k$-dimensional D-Marker feature vector, \(\mathbf{D_i} = \{\mathbf{d_1}, \mathbf{d_2}, \ldots, \mathbf{d_k}\}\) where $k=225$ in our implementation. 

The empirical validation of these D-Marker features was demonstrated by Dibeklioğlu \textit{et al.} \cite{dibekliouglu2015recognition}, establishing their discriminative power for genuine smile recognition. Our ablation analysis in Table \ref{tab:comprehensive_ablation} demonstrates the individual contributions of Duration, Position, and Motion feature categories, confirming that each feature group captures essential aspects of genuine smile dynamics and contributes meaningfully to classification performance. This hand-crafted D-Marker representation encodes the geometric and temporal patterns of genuine smiles in a structured, low-dimensional feature space, offering complementary information to data-driven embeddings for effective feature-level integration in our proposed framework.

\subsection{HadaSmileNet Architecture}\label{sec:dfusesmilenet}

\begin{figure*}
    \centering
    \includegraphics[width=1\linewidth]{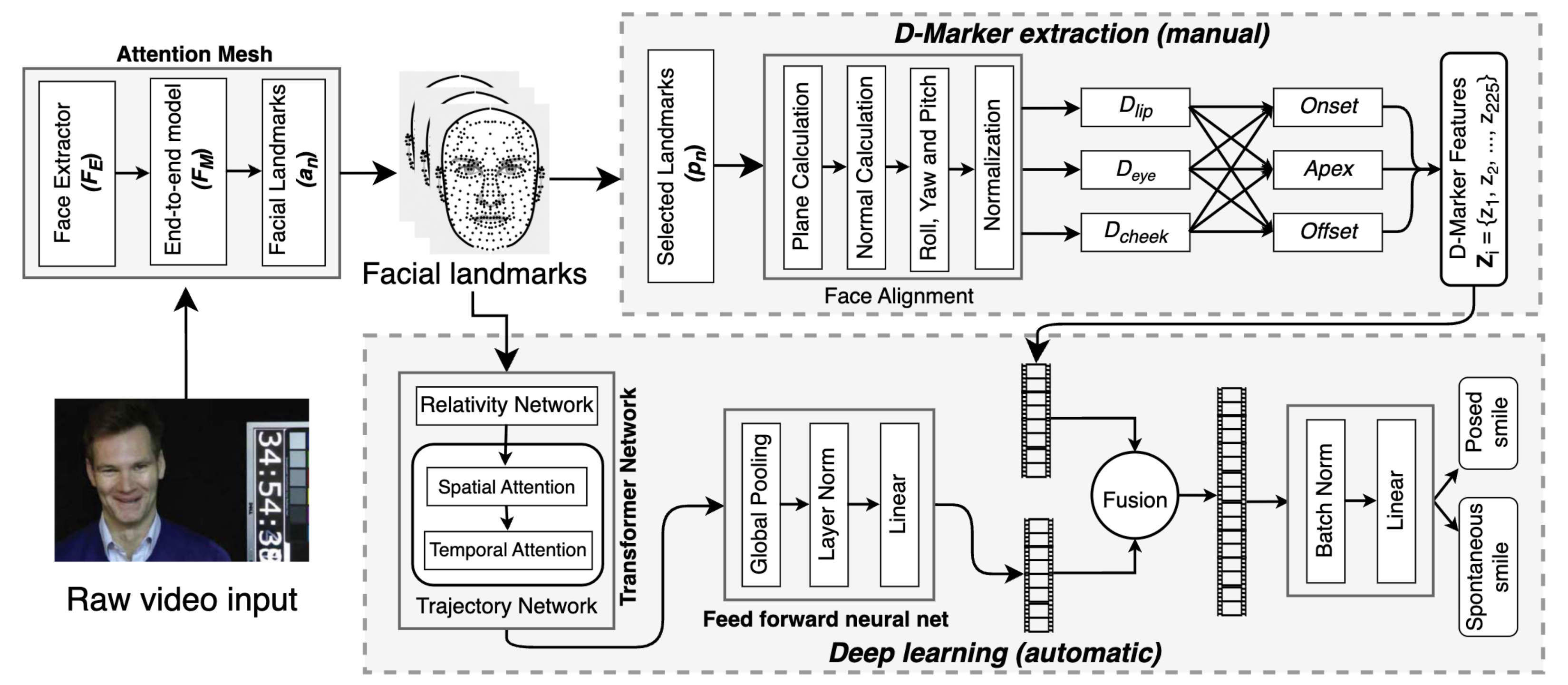}
    \caption{Overview of our proposed framework for genuine smile recognition. The pipeline consists of two parallel streams: (1) an automatic feature extraction path using Attention Mesh for facial landmark detection followed by transformer-based feature learning, and (2) a D-Marker extraction path that computes handcrafted physiological features. These complementary features are combined through various fusion mechanisms to achieve robust smile classification.}
    \label{fig:archi}
\end{figure*}

The architecture diagram of the proposed method is illustrated in Figure \ref{fig:archi}. The architecture mainly has two parts: the D-Marker extraction using manual handcrafted approaches, as detailed in the previous section, and the automatic deep-learning part.

\noindent\textbf{\textit{Automatic Feature Extraction.}} To obtain robust automatic representations of facial dynamics, we adopt the current state-of-the-art transformer-based architecture, MeshSmileNet \cite{faroque2022less}. This architecture serves as our feature extractor $\mathcal{F}$ that processes temporal sequences of 3D facial landmarks to produce discriminative representations.

Given the normalized facial landmarks from Section \ref{sec:handcrafted_features}, we construct input sequences $\mathbf{X}_i = \{\mathbf{x}_1, \mathbf{x}_2, \ldots, \mathbf{x}_{T_i}\}$ where each frame $\mathbf{x}_t \in \mathbb{R}^{478 \times 3}$ contains the 3D coordinates of all 478 facial landmarks detected by AttentionMesh. To handle variable-length sequences, we apply temporal normalization by either truncating longer sequences or zero-padding shorter ones to a fixed length $T = 64$ frames.

The MeshSmileNet architecture consists of three primary components that collectively learn spatiotemporal representations: \\\textit{(1) Relativity Network:} This component analyzes the spatial geometric relationships among facial landmarks within each frame. It employs CurveNet blocks \cite{xiang2021walk} to aggregate landmarks into meaningful curves based on their geometric proximity and functional relationships. For each frame $\mathbf{x}_t$, the relativity network produces spatial embeddings $\mathbf{s}_t \in \mathbb{R}^{d_s}$ that capture the geometric configuration of facial features:
\begin{equation}
    \mathbf{s}_t = \text{CurveNet}(\mathbf{x}_t),
\end{equation}
where $d_s = 128$ is the spatial embedding dimension.\\\textit{(2) Trajectory Network:} This component tracks the temporal evolution of landmark movements across the sequence. It utilizes a multi-head self-attention mechanism to model long-range dependencies and temporal dynamics. The trajectory network processes the sequence of spatial embeddings $\{\mathbf{s}_1, \mathbf{s}_2, \ldots, \mathbf{s}_T\}$ to produce temporal features:
\begin{equation}
    \mathbf{T}_i = \text{MultiHeadAttention}(\mathbf{S}_i, \mathbf{S}_i, \mathbf{S}_i),
\end{equation}
where $\mathbf{S}_i = [\mathbf{s}_1; \mathbf{s}_2; \ldots; \mathbf{s}_T] \in \mathbb{R}^{T \times d_s}$ is the concatenated spatial embeddings, and $\mathbf{T}_i \in \mathbb{R}^{T \times d_s}$ captures temporal dependencies.\\\textit{(3) Feature Aggregation:} The final video-level representation $\mathbf{H}_i$ is obtained by applying global average pooling over the temporal dimension, followed by a linear projection:
\begin{equation}
    \mathbf{H}_i = \mathbf{W}_h \cdot \frac{1}{T} \sum_{t=1}^{T} \mathbf{T}_i[t, :] + \mathbf{b}_h,
\end{equation}
where $\mathbf{W}_h \in \mathbb{R}^{D \times d_s}$ and $\mathbf{b}_h \in \mathbb{R}^{D}$ are learned parameters, and $D = 256$ is the dimension of the final learned representation.

\noindent\textbf{\textit{Feature Fusion Mechanism.}} The core innovation of HadaSmileNet lies in its direct fusion of learned transformer features $\mathbf{H}_i \in \mathbb{R}^{D}$ with handcrafted D-Marker features $\mathbf{Z}_i \in \mathbb{R}^{M}$. We systematically evaluate 15 different fusion strategies, ranging from simple concatenation to complex attention-based mechanisms. Through comprehensive empirical analysis, we identify Hadamard multiplicative fusion as the optimal strategy.

To enable element-wise multiplication, we first align the dimensionalities of both feature vectors through learned linear transformations:
\begin{align}
    \mathbf{H}_i^* &= \mathbf{W}_H \mathbf{H}_i + \mathbf{b}_H, \\
    \mathbf{Z}_i^* &= \mathbf{W}_Z \mathbf{Z}_i + \mathbf{b}_Z,
\end{align}
where $\mathbf{W}_H \in \mathbb{R}^{Q \times D}$, $\mathbf{W}_Z \in \mathbb{R}^{Q \times M}$, and $\mathbf{b}_H, \mathbf{b}_Z \in \mathbb{R}^{Q}$ are learnable parameters that project both representations to a common dimension $Q = 128$.

The Hadamard fusion operation is then defined as:
\begin{equation}
    \mathbf{F}_i = \mathbf{H}_i^* \odot \mathbf{Z}_i^*,
\end{equation}
where $\odot$ denotes element-wise multiplication. This fusion strategy enables direct multiplicative interactions between corresponding dimensions of the learned and handcrafted features, allowing the model to amplify or suppress feature components based on their complementary relationships.

The effectiveness of Hadamard fusion stems from its ability to create multiplicative gates that modulate the learned features based on domain-specific D-Marker cues. Unlike additive fusion (concatenation) that simply combines features linearly, or attention-based fusion that requires additional parameters, Hadamard fusion creates non-linear interactions while maintaining parameter efficiency. The multiplicative nature enables the D-Marker features to act as selective gates, emphasizing relevant learned features while suppressing noise or irrelevant patterns.

\noindent\textbf{\textit{Classification Network.}} The fused representation $\mathbf{F}_i$ is passed through a lightweight classification head to produce the final smile prediction. The classifier employs layer normalization for stable training followed by a single linear transformation:
\begin{align}
    \mathbf{F}_i^{norm} &= \text{LayerNorm}(\mathbf{F}_i), \\
    \hat{y}_i &= \sigma(\mathbf{W}_c \mathbf{F}_i^{norm} + \mathbf{b}_c),
\end{align}
where $\mathbf{W}_c \in \mathbb{R}^{1 \times 256}$, $\mathbf{b}_c \in \mathbb{R}$ are the classifier parameters, $\sigma$ is the sigmoid activation function, and $\hat{y}_i \in [0,1]$ represents the predicted probability of the smile being genuine.

The simplified classification architecture, compared to the multi-head design in multi-task learning approaches, reduces computational overhead while maintaining discriminative power. The layer normalization ensures stable gradients and consistent feature magnitudes, while the single linear layer provides sufficient capacity for the binary classification task given the rich, pre-processed fusion features.

\noindent\textbf{\textit{Training Procedure.}} Unlike multi-task learning approaches that require careful balancing of multiple loss terms, our framework employs a single binary cross-entropy loss (Equation 4) for end-to-end training. The entire network, including the transformer backbone \(\mathcal{F}\), projection layers \(\mathbf{W}_H\) and \(\mathbf{W}_Z\), and classifier, is trained jointly using the Adam optimizer with an initial learning rate of \(1 \times 10^{-4}\) and cosine annealing schedule. During training, both D-Marker features and video sequences are utilized to enable the model to learn optimal fusion representations. The training process eliminates the computational overhead and hyperparameter tuning complexity associated with multi-task frameworks while achieving superior discriminative performance. The key architectural advantage of HadaSmileNet over previous multi-task approaches is its simplicity and efficiency. By eliminating the auxiliary D-Marker prediction head (which contains 56,024 parameters including a 256×216 linear layer and associated LayerNorm), our method reduces the total parameter count while achieving superior performance. Direct feature fusion ensures that domain knowledge is explicitly integrated into the learned representations, creating more discriminative and interpretable features for genuine smile recognition. 
\begin{figure*}[!t]
    \centering
    \includegraphics[width=0.9\linewidth]{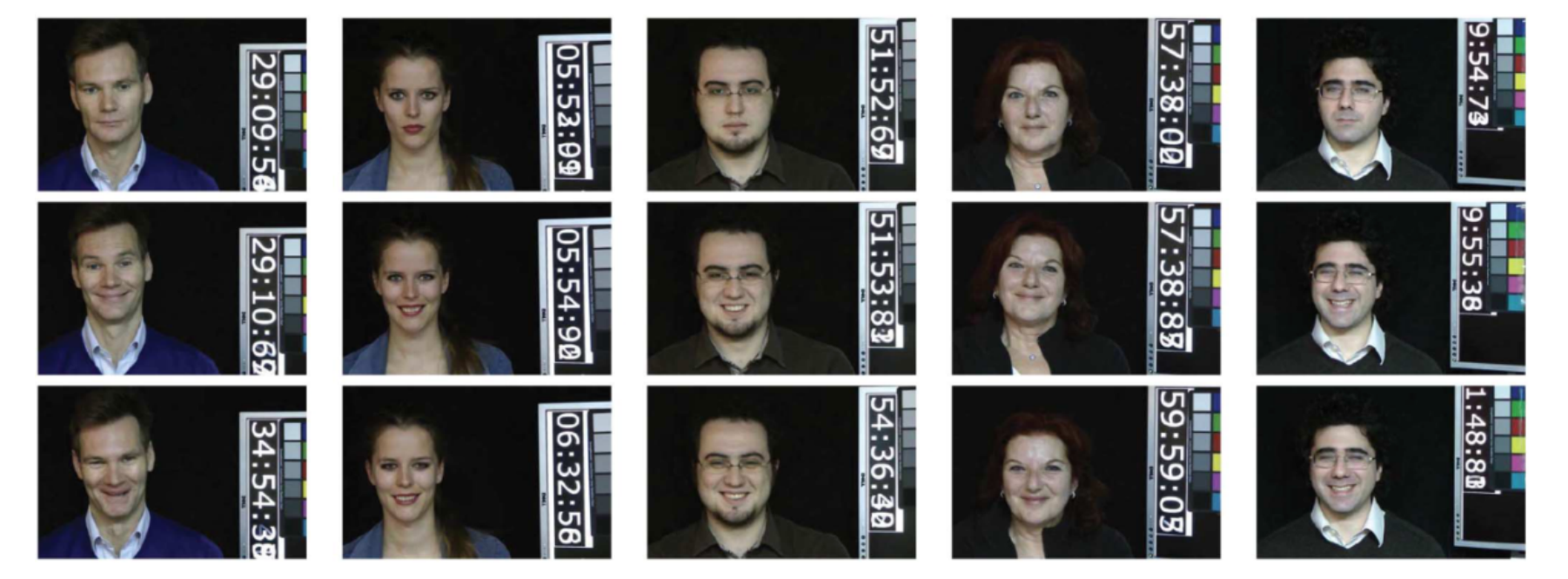}
    \caption{Data samples randomly drawn from the UvA-NEMO database showing neutral face (top), posed enjoyment smile (middle), and genuine enjoyment smile (bottom).}
    \label{fig:samples}
\end{figure*}

\vspace*{-1em}

\noindent\textbf{\textit{Inference Procedure.}} A core advantage of the proposed framework lies in its inference strategy, which eliminates the dependency on D-Marker features during testing. Unlike the training phase where both video sequences and handcrafted D-Markers are required, inference operates exclusively on video input, requiring only the transformer backbone \(\mathcal{F}\) and the trained classifier \(C\). The learned projection layers \(\mathbf{W}_H\) and \(\mathbf{W}_Z\) effectively encode the complementary information from D-Markers into the network weights during training, enabling the model to implicitly leverage this domain knowledge without explicit feature computation at test time. This design choice serves multiple objectives: \textbf{(1)} \textit{Computational Efficiency}: Eliminates the computational overhead of landmark detection, geometric normalization, and D-Marker computation during inference, reducing processing time. \textbf{(2)} \textit{Practical Deployment}: Enables real-time processing in resource-constrained environments where manual feature extraction would be prohibitive. \textbf{(3)} \textit{Fair Comparison}: Ensures direct comparability with existing deep learning methods that operate solely on visual input during testing. \textbf{(4)} \textit{Robustness}: Reduces susceptibility to landmark detection failures or geometric estimation errors that could compromise D-Marker quality in challenging conditions.

\section{Experiments}
\subsection{Setup}


\noindent\textbf{\textit{Datasets.}} We evaluate our proposed method independently on four widely-used benchmark datasets: \textbf{(i)} The UvA-NEMO smile database \cite{dibekliouglu2015recognition} is the most comprehensive dataset for smile authenticity analysis, comprising 1240 high-definition videos (597 genuine, 643 posed) recorded at 1920×1080 resolution and 50 FPS. The dataset features 400 subjects (185 female, 215 male) across a broad age spectrum (8-76 years), with controlled illumination conditions enabling focus on expression dynamics. \textbf{(ii)} The BBC dataset presents unique challenges through its collection of celebrity interviews, containing 20 smile videos (10 genuine, 10 posed) recorded in real-world conditions. The lower resolution (314×286) and varying illumination make it particularly suitable for evaluating robustness. \textbf{(iii)} The MMI facial expression dataset contributes 187 smile videos (138 genuine at 640×480/29 FPS, 49 posed at 720×576/25 FPS), offering diversity in recording conditions. \textbf{(iv)} The SPOS dataset \cite{pfister2011differentiating} includes grayscale sequences captured at 640×480 resolution and 25 FPS. We utilize its gray-scale smile sequences, which comprise both genuine and posed expressions under controlled settings.
These datasets present varying challenges through their different recording conditions, resolutions, and subject demographics, enabling comprehensive evaluation of our method's generalization capabilities. Notably, while UvA-NEMO offers ideal conditions for analyzing subtle expression differences, the BBC dataset tests robustness to real-world variations, and MMI and SPOS provide additional validation across different video qualities and frame rates. The dataset details are summarized in Table \ref{tab:dataset}. Some data samples randomly selected from the UvA-NEMO database can be visualized in Figure \ref{fig:samples}. 
\begin{table}[!t]
\centering
\renewcommand{\arraystretch}{1.1}
\scriptsize
\caption{Details of the benchmark smile datasets used.}
\label{tab:dataset}
\adjustbox{width=\columnwidth,center}
{
\begin{tabular}{lccccccc}
\toprule
& \multicolumn{2}{c}{\textbf{Video Spec.}} & \multicolumn{2}{c}{\textbf{Number of Videos}} & \multicolumn{2}{c}{\textbf{Number of Subjects}} \\
\cmidrule(lr){2-3} \cmidrule(lr){4-5} \cmidrule(lr){6-7}
\textbf{Dataset} & \textbf{Resolution} & \textbf{FPS} & \textbf{Genuine} & \textbf{Posed} & \textbf{Genuine} & \textbf{Posed} \\
\midrule
UvA-NEMO & 1920×1080 & 50 & 597 & 643 & 357 & 368 \\
\midrule
BBC & 314×286 & 25 & 10 & 10 & 10 & 10 \\
\midrule
MMI & \makecell{720×576\\640×480} & \makecell{29\\25} & 138 & 49 & 9 & 25 \\
\midrule
SPOS & 640×480 & 25 & 66 & 14 & 7 & 7 \\
\bottomrule
\end{tabular}
}
\end{table}

\noindent\textbf{\textit{Evaluation Protocol.}} We adopt rigorous cross-validation protocols for comprehensive evaluation across datasets. For UvA-NEMO, we employ 10-fold cross-validation following established protocols \cite{dibekliouglu2015recognition}. The BBC, MMI, and SPOS datasets are evaluated using 10-fold, 9-fold, and 7-fold cross-validation respectively, ensuring fair comparison with previous works \cite{wu2014spontaneous}. To ensure robust performance estimation, we conduct ten independent runs for each dataset, with each subset serving as test data exactly once. Special care is taken to maintain subject independence across train-test splits, preventing potential data leakage. Performance is measured using \textbf{classification accuracy} averaged across all folds.

\noindent\textbf{\textit{Implementation Details\footnote{Codes and models are available at: \href{https://github.com/junayed-hasan/smile-recognition-fusion}{https://github.com/junayed-hasan/smile-recognition-fusion}.}.}} Our framework is implemented in PyTorch and trained on an NVIDIA Tesla T4 GPU on the Amazon AWS EC2 server. The preprocessing pipeline extracts 478 3D facial landmarks using Attention Mesh, which are subsequently transformed into fixed-length sequences of 16 frames for consistent input dimensionality. The network architecture consists of a temporal model with a CurveNet encoder, followed by our proposed fusion module that combines transformer features with D-Marker representations. Training proceeds for 300 epochs with a batch size of 16, using the AdamW optimizer with an initial learning rate of 5e-4. The spatial-temporal transformer employs 6 blocks for spatial attention and 3 blocks for temporal modeling, each with 4 attention heads. We utilize binary cross-entropy loss for smile classification, with all network weights initialized using He initialization. Layer normalization and dropout (p=0.1) are applied throughout the network to prevent overfitting.

\begin{table}[!t]

\renewcommand{\arraystretch}{1}
\centering
\small
\setstretch{1}
\caption{Performance comparison with state-of-the-art methods on genuine smile recognition. Results reported as classification accuracy (\%). Deep learning-based methods are highlighted in \colorbox{gray!20}{gray}. Best results are in \textbf{bold}, second-best are \underline{underlined}. $^{***}$p-value$<$0.001 vs. previous best DL method.}

\label{tab:benchmark}
\begin{tabular}{lcccc}
\toprule
Method & UvA- & MMI & SPOS & BBC \\
& NEMO & & & \\
\midrule
Cohn'04 \cite{cohn2004timing} & 77.3 & 81.0 & 73.0 & 75.0 \\
Dibeklioglu'10 \cite{dibeklioglu2010eyes} & 71.1 & 74.0 & 68.0 & 85.0 \\
Pfister'11 \cite{pfister2011differentiating} & 73.1 & 81.0 & 67.5 & 70.0 \\
Wu'14 \cite{wu2014spontaneous} & \underline{91.4} & 86.1 & 79.5 & 90.0 \\
Dibeklioglu'15 \cite{dibekliouglu2015recognition} & 89.8 & 88.1 & 77.5 & 90.0 \\
Wu'17 \cite{wu2017spontaneous} & \textbf{93.9} & 92.2 & 81.2 & 90.0 \\
Mandal'17 \cite{mandal2017spontaneous} & 80.4 & - & - & - \\
\rowcolor{gray!20} Mandal'16 \cite{mandal2017distinguishing} & 78.1 & - & - & - \\
\rowcolor{gray!20} RealSmileNet'20 \cite{yang2020realsmilenet} & 82.1 & 92.0 & 86.2 & 90.0 \\
\rowcolor{gray!20} PSTNet'22 \cite{fan2022pstnet} & 72.9 & 94.3 & 87.1 & \underline{95.0} \\
\rowcolor{gray!20} P4Transformer'21 \cite{fan2021point} & 74.9 & 91.3 & 82.9 & 85.0 \\
\rowcolor{gray!20} Vanilla ViT'20 \cite{dosovitskiy2020image} & 78.4 & \underline{99.0} & 93.5 & \underline{95.0} \\
\rowcolor{gray!20} MeshSmileNet'22 \cite{faroque2022less} & 85.0 & \underline{99.0} & 94.4 & \underline{95.0} \\
\rowcolor{gray!20} DeepMarkerNet'24 \cite{HASAN2024148} & 87.9 & \textbf{99.7} & \underline{97.8} & \underline{95.0} \\
\midrule
Ours (HadaSmileNet) & 88.7$^{***}$ & \textbf{99.7} & \textbf{98.5}$^{***}$ & \textbf{100.0}$^{***}$ \\
\bottomrule
\end{tabular}
\end{table}

\subsection{Comparison with state-of-the-art}

Table \ref{tab:benchmark} presents the empirical results achieved for our method versus all existing methods in the literature. From the table, one may observe the following:  \textbf{\textit{(i)}} We achieve state-of-the-art performance on three out of four datasets, with statistically significant gains (p-value $<$ 0.001) on SPOS (98.5\%) and BBC (100\%), surpassing both traditional feature-based approaches and recent deep learning methods. \textbf{\textit{(ii)}} While Wu \textit{et al.} \cite{wu2017spontaneous} maintains the lead on UvA-NEMO (93.9\% vs. our 88.7\%), their method relies on manual landmark initialization and extensive preprocessing, whereas our approach is fully automatic and end-to-end trainable. \textbf{\textit{(iii)}} Notably, we improve upon DeepMarkerNet \cite{HASAN2024148}, the previous state-of-the-art that also utilizes D-Marker information but through multi-task learning rather than direct feature fusion. Our method shows consistent gains across all datasets (UvA-NEMO: +0.8\%, MMI: +0.7\%, SPOS: +1.3\%, BBC: +5.0\%), demonstrating the superiority of explicit feature interaction over auxiliary task supervision.

\begin{figure*}[t]
\centering
\includegraphics[width=0.9\textwidth]{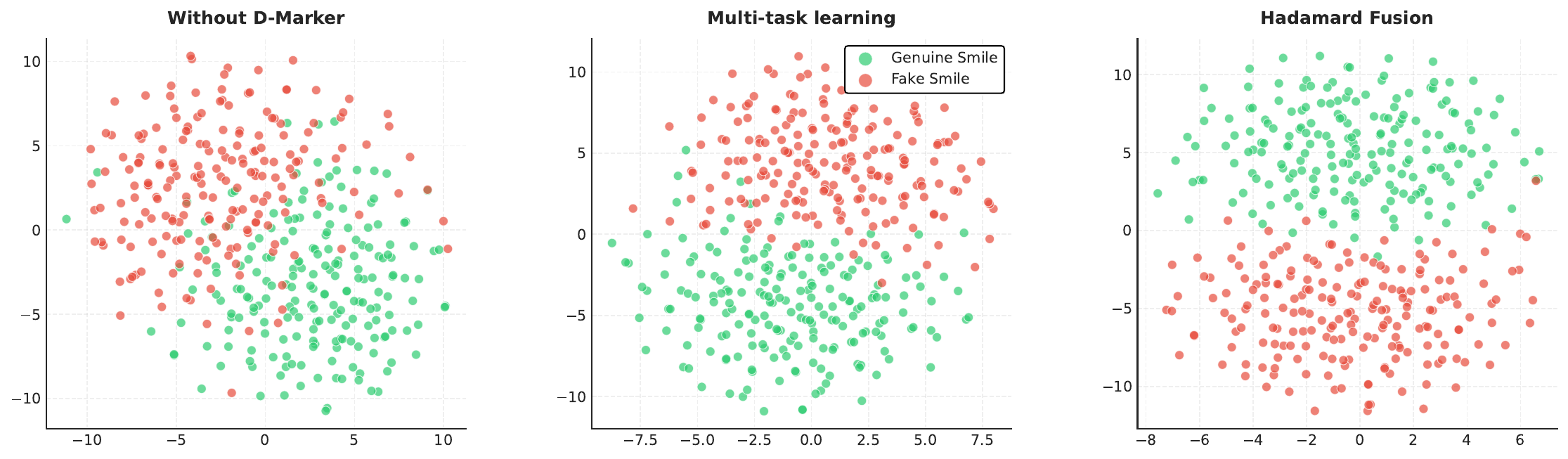}
\caption{t-SNE visualization of learned feature embeddings on the UvA-NEMO dataset. Left: baseline without D-Marker features shows significant class overlap. Middle: DeepMarkerNet's multi-task learning \cite{HASAN2024148} demonstrates improved but still overlapping separation. Right: Our Hadamard fusion achieves better class separation with clearer decision boundary.}
\label{fig:tsne}
\end{figure*}

The effectiveness of our Hadamard fusion strategy is further visually demonstrated through the t-SNE visualizations in Figure \ref{fig:tsne}. The baseline approach without D-Marker features shows significant overlap between genuine and posed smiles. While DeepMarkerNet's multi-task learning approach improves feature separability, decision boundaries remain overlapping. In contrast, our Hadamard fusion method achieves clear class separation, demonstrating that direct multiplicative interactions create more discriminative representations compared to indirect supervision through auxiliary tasks.

\begin{table}[t]
\centering

\footnotesize
\caption{Performance comparison of different feature fusion techniques across benchmark datasets. The \textbf{bold} values indicate the best performance for each dataset, while \underline{underlined} values represent the second-best performance.}
\label{tab:fusion_results}

\resizebox{\columnwidth}{!}{
\begin{tabular}{lccccc}
\toprule
Fusion Technique & UvA- & MMI & SPOS & BBC \\
& NEMO & & & \\
\midrule
Concatenation & \underline{87.9} & \textbf{99.7} & 97.2 & 95.0 \\
Gated Concatenation & \underline{87.9} & 94.7 & 93.0 & 95.0 \\
Additive Fusion & 81.4 & 89.7 & 92.0 & 85.0 \\
Hadamard Fusion & \textbf{88.7} & \textbf{99.7} & \textbf{98.5} & \textbf{100.0} \\
Gated + Hadamard Fusion & \underline{87.9} & \textbf{99.7} & 93.4 & \textbf{100.0} \\
\midrule
Attention & 87.1 & 89.4 & 91.6 & 90.0 \\
Multi-head Attention & 85.4 & 91.7 & 90.4 & 85.0 \\
Cross Attention & 83.8 & 92.3 & 85.7 & 90.0 \\
Multi-head Cross Attention & 85.4 & 92.1 & 86.2 & 85.0 \\
\midrule
Feature-wise Linear Modulation (FiLM) & 83.8 & 94.7 & 92.8 & 90.0 \\
FiLM + Hadamard & 87.1 & 92.1 & 97.8 & 95.0 \\
\midrule
Bilinear Pooling & 83.8 & 95.6 & 92.8 & 95.0 \\
Bilinear + Hadamard & 87.1 & 93.1 & 96.7 & \textbf{100.0} \\
Factorized Bilinear & 83.8 & 93.3 & 97.7 & 95.0 \\
Factorized + Hadamard & 85.4 & 94.7 & 95.4 & \textbf{100.0} \\
\bottomrule
\end{tabular}}

\end{table}

\subsection{Ablation studies}
\noindent{\textbf{Ablation on fusion techniques.}} We systematically evaluate 15 feature fusion strategies to validate our choice of Hadamard fusion. Table \ref{tab:fusion_results} presents classification accuracy across datasets. The results demonstrate several critical insights: \textbf{\textit{(i)}} Hadamard multiplicative fusion achieves optimal performance across all datasets, effectively preserving spatial correspondence between transformer and D-Marker features while enabling element-wise feature interactions. \textbf{\textit{(ii)}} Despite their theoretical sophistication, attention-based mechanisms exhibit suboptimal performance, indicating that the complementary nature of transformer and D-Marker features is better captured through direct multiplicative interactions rather than learned attention weights. \textbf{\textit{(iii)}} The integration of Hadamard multiplication consistently enhances other fusion methods, with notable improvements observed in FiLM and Bilinear Pooling variants, suggesting its fundamental value in multimodal feature integration. \textbf{\textit{(iv)}} Simple concatenation achieves competitive performance, particularly on MMI (99.7\%), demonstrating that allowing networks to learn feature interactions through subsequent layers can be effective when sufficient training data is available.

\noindent{\textbf{Ablation on D-Marker feature components.}} We also evaluate the contribution of duration, position, and motion features extracted from facial dynamics across three regions. Table \ref{tab:comprehensive_ablation} shows that duration features provide the most discriminative information for genuine smile recognition. The removal of duration features causes substantial performance degradation (UvA-NEMO: -5.5\%), confirming that temporal dynamics of smile evolution constitute the strongest authenticity cues. Motion and position features contribute complementarily, with motion features demonstrating higher importance on challenging datasets. This hierarchical importance validates the physiological foundation of D-Markers, where timing patterns distinguish genuine emotional expressions from deliberate facial movements. The consistent degradation pattern indicates that all three feature categories capture essential aspects of genuine smile dynamics.

\noindent{\textbf{Ablation on facial regions.}} We investigate the contribution of D-Marker features extracted from lips, eyes, and cheeks through systematic exclusion experiments. Table \ref{tab:comprehensive_ablation} illustrates the performance impact when removing features from individual facial regions. Consistent with physiological understanding of Duchenne smiles, cheek elevation features demonstrate the highest discriminative power, with their exclusion resulting in substantial performance drops across all datasets. This aligns with the critical role of zygomaticus major muscle activation in genuine smiles. Eye region features, capturing the characteristic crinkling around orbicularis oculi muscles, contribute significantly to classification accuracy, particularly evident in the UvA-NEMO dataset performance decline. Lip dynamics show the smallest individual contribution to authenticity classification, suggesting that subtle differences between genuine and posed smiles are more pronounced in upper facial regions where involuntary muscle contractions occur during genuine emotional expressions.

\noindent{\textbf{Ablation on landmark selection.}} We validate our landmark selection methodology by systematically replacing our physiologically-motivated 11 key points with their nearest neighbors in the AttentionMesh 478-point topology. Table \ref{tab:comprehensive_ablation} demonstrates consistent performance degradation as landmarks deviate from anatomically relevant positions. The gradient decline pattern (2.3\%, 4.8\%, 6.1\% drops for first, second, third nearest replacements respectively) confirms that our selection strategy successfully identifies anatomically relevant points that capture the most discriminative muscle activation patterns. Eye corner landmarks exhibit the highest sensitivity to positional accuracy, emphasizing the importance of precise localization for capturing subtle periocular muscle movements characteristic of genuine smiles. This validates our precision-based selection approach over random or convenience-based landmark choices.

\noindent{\textbf{Ablation on transformer architectures.}} We evaluate the generalizability of our Hadamard fusion approach across different transformer-based backbones to demonstrate architectural independence. Table \ref{tab:comprehensive_ablation} compares performance when integrating our fusion framework with different architectures. The consistent improvements across architectures (+3.2\% for RealSmileNet, +3.7\% for MeshSmileNet) validate the effectiveness of our direct feature fusion paradigm. MeshSmileNet demonstrates the strongest compatibility with D-Marker features, likely due to its specialized curve-based spatial encoding that aligns well with the geometric nature of D-Marker computations. The robust performance gains confirm that benefits stem from the complementary information provided by D-Markers rather than architecture-specific optimizations, establishing broad applicability for transformer-based smile recognition systems.

\begin{table}[t]
\centering
\caption{Comprehensive ablation analysis on UvA-NEMO dataset. \checkmark indicates inclusion, \ding{55} indicates exclusion.}
\label{tab:comprehensive_ablation}
\adjustbox{width=\columnwidth,center}
{
\begin{tabular}{lcccc}
\toprule
\textbf{Component} & \textbf{Duration} & \textbf{Position} & \textbf{Motion} & \textbf{Acc (\%)} \\
\midrule
\multirow{4}{*}{{\textit{D-Marker}}} 
& \ding{55} & \checkmark & \checkmark & 83.2 \\
& \checkmark & \ding{55} & \checkmark & 85.4 \\
& \checkmark & \checkmark & \ding{55} & 84.9 \\
& \checkmark & \checkmark & \checkmark & \textbf{88.7} \\
\midrule
\textbf{Component} & \textbf{Lip} & \textbf{Eye} & \textbf{Cheek} & \textbf{Acc (\%)} \\
\midrule
\multirow{4}{*}{{\textit{Regions}}} 
& \ding{55} & \checkmark & \checkmark & 87.6 \\
& \checkmark & \ding{55} & \checkmark & 84.5 \\
& \checkmark & \checkmark & \ding{55} & 81.9 \\
& \checkmark & \checkmark & \checkmark & \textbf{88.7} \\
\midrule
\textbf{Configuration} & \textbf{Original} & \textbf{1st Nearest} & \textbf{2nd/3rd Nearest} & \textbf{Acc (\%)} \\
\midrule
\multirow{4}{*}{{\textit{Landmarks}}} 
& \checkmark & \ding{55} & \ding{55} & \textbf{88.7} \\
& \ding{55} & \checkmark & \ding{55} & 86.4 \\
& \ding{55} & \ding{55} & 2nd: \checkmark & 83.9 \\
& \ding{55} & \ding{55} & 3rd: \checkmark & 82.6 \\
\midrule
\textbf{Architecture} & \textbf{RealSmileNet \cite{yang2020realsmilenet}} & \textbf{MeshSmileNet \cite{faroque2022less}} & \textbf{Fusion} & \textbf{Acc (\%)} \\
\midrule
\multirow{4}{*}{{\textit{Backbone}}} 
& \checkmark & \ding{55} & \ding{55} & 82.1 \\
& \checkmark & \ding{55} & \checkmark & 85.3 \\
& \ding{55} & \checkmark & \ding{55} & 85.0 \\
& \ding{55} & \checkmark & \checkmark & \textbf{88.7} \\
\bottomrule
\end{tabular}
}

\end{table}

\section{Discussion}

\noindent\textbf{Cross-dataset experiments.} 
To evaluate generalization capabilities, cross-dataset experiments were conducted that compared HadaSmileNet with recent deep learning methods. Training in UvA-NEMO and testing on BBC, MMI, and SPOS datasets under cross-domain conditions revealed a consistent superiority of the proposed method. Table \ref{tab:cross-data} demonstrates that direct feature fusion enables superior generalization compared to auxiliary task supervision, achieving 1.5\%, 0.2\%, and 1.4\% improvements over DeepMarkerNet on BBC, MMI, and SPOS, respectively. These results validate the hypothesis that multiplicative feature interactions create more transferable representations that are effective in generalizing across diverse recording conditions and demographic variations.

\begin{figure}[t]
\centering
\includegraphics[width=\columnwidth]{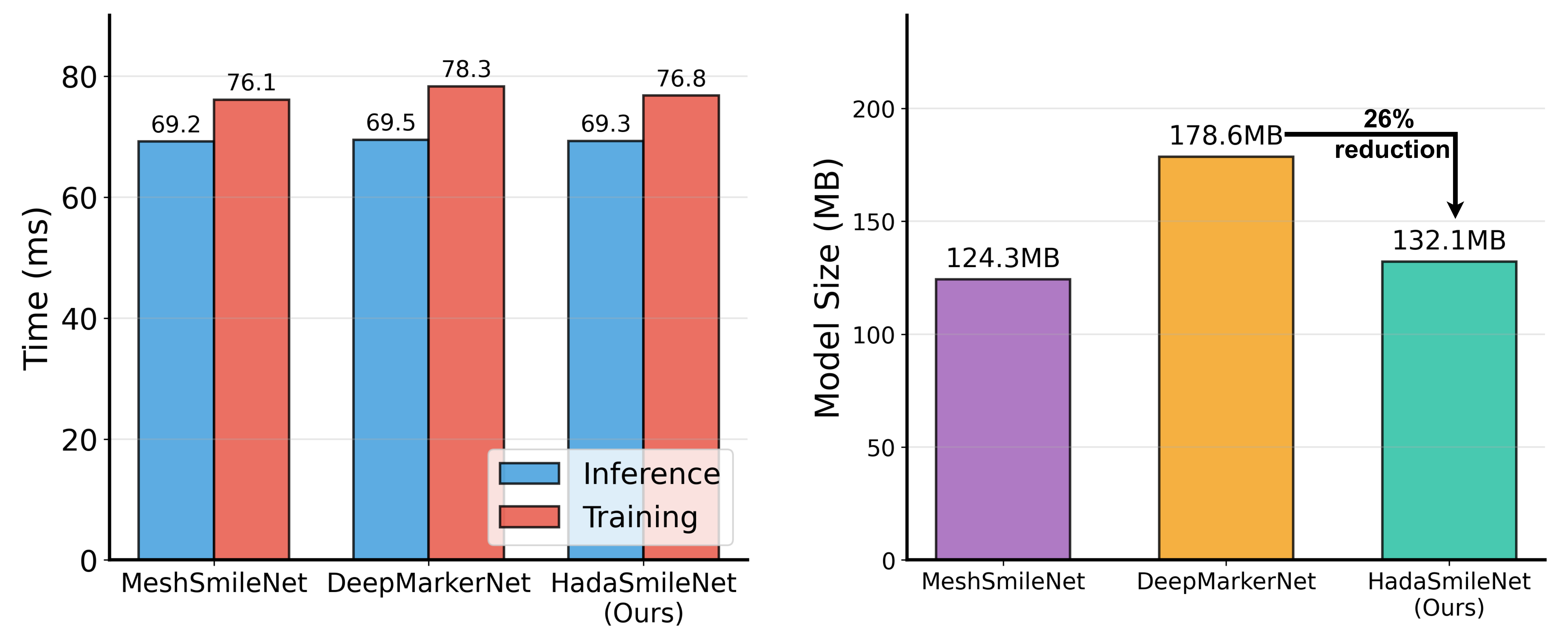}
\caption{Computational efficiency analysis showing reduced training, inference time, and a 26\% parameter reduction in our proposed method compared to current state-of-the-art methods.}
\label{fig:computational-analysis}
\end{figure}

\noindent\textbf{Computational cost analysis.} 
Comprehensive timing experiments measuring inference, training costs, and model sizes were carried out on an Amazon Linux AMI operating system with NVIDIA T4 14GB Tensor Core GPU on the AWS EC2 cloud server. Figure \ref{fig:computational-analysis} reveals significant advantages of the fusion approach, most notably a substantial 26\% parameter reduction and 42.3\% training time reduction compared to DeepMarkerNet (both p-value $<$ 0.001). The method achieves an inference time comparable to MeshSmileNet and DeepMarkerNet, demonstrating minimal deployment overhead. Training efficiency significantly outperforms DeepMarkerNet by eliminating auxiliary head computations and complex loss balancing. This parameter reduction translates to reduced memory requirements, faster loading times, and improved deployment feasibility in resource-constrained environments.

\begin{table}[t]
\centering
\caption{Cross-dataset experiments using UvA-NEMO for training.}
\label{tab:cross-data}
\begin{tabular}{lccc}
\toprule
\textbf{Method} & \textbf{BBC} & \textbf{MMI} & \textbf{SPOS} \\
\midrule
RealSmileNet \cite{yang2020realsmilenet} & 80.0 & 92.5 & 82.5 \\
MeshSmileNet \cite{faroque2022less} & 86.0 & 99.2 & 90.3 \\
DeepMarkerNet \cite{HASAN2024148} & 88.5 & 99.4 & 93.8 \\
\textbf{HadaSmileNet (Ours)} & \textbf{90.0} & \textbf{99.6} & \textbf{95.2} \\
\bottomrule
\end{tabular}
\end{table}

\noindent\textbf{Feature fusion vs. multi-task learning.} 
The direct feature fusion approach offers fundamental advantages over multi-task frameworks through architectural simplicity and training stability. Unlike DeepMarkerNet's complex loss weighting and competing gradients between classification and regression tasks, Hadamard fusion enables direct multiplicative interaction without hyperparameter tuning complexity. The unified gradient flow focused solely on classification objectives prevents gradient conflicts that plague multi-task approaches. The fusion mechanism automatically learns optimal feature combinations rather than relying on manually tuned loss weights, while t-SNE visualizations (Figure \ref{fig:tsne}) demonstrate superior discriminative feature representations with enhanced class separation compared to auxiliary task supervision.

\noindent\textbf{Fusion-based domain knowledge integration.} 
The approach represents a paradigm shift in the integration of physiological domain knowledge through direct fusion of features rather than auxiliary supervision. Hadamard multiplicative fusion creates meaningful interactions between transformer representations and physiologically-grounded D-Markers, allowing selective amplification of learned features based on domain-specific facial muscle activation cues. A comprehensive evaluation of 15 fusion strategies validates that multiplicative interaction outperforms sophisticated attention mechanisms and bilinear pooling. The framework enables transformers to learn representations naturally aligned with physiological understanding, with D-Markers acting as multiplicative gates that modulate feature importance based on muscle activation patterns.

\noindent\textbf{Deployment and ethical considerations.} 
Real-world deployment of HadaSmileNet requires integration with complementary technologies including video acquisition systems, facial detection pipelines, and decision support frameworks for human-computer interaction applications. The framework demonstrates compatibility with existing computer vision infrastructures through its transformer-based architecture and standard input formats. However, deployment raises ethical considerations regarding facial data privacy and algorithmic fairness. The system processes sensitive biometric information requiring robust data protection protocols, informed consent mechanisms, and compliance with privacy regulations. Fairness concerns emerge from potential demographic biases in training data and the exclusion of individuals with facial differences or neurodevelopmental conditions whose D-Marker patterns may deviate from normative assumptions, necessitating inclusive dataset curation and bias mitigation strategies.

\noindent\textbf{Limitations.} 
Several limitations warrant consideration regarding performance and inclusivity. The method could not surpass traditional feature-based approaches in UvA-NEMO that benefit from manual landmark initialization. Training requires D-Marker extraction, adding computational overhead compared to end-to-end approaches. Dataset limitations include reliance on acted expressions in benchmark datasets, where approximately half of UvA-NEMO samples represent posed rather than spontaneous emotions, potentially affecting the ecological validity of learned representations. The framework excludes individuals with facial differences including cleft lip, Down syndrome, or other neurodevelopmental conditions whose D-Marker patterns deviate from normative facial muscle activation assumptions, raising concerns about algorithmic fairness and generalizability across diverse populations. The dependency on accurate facial landmark detection can compromise performance in occlusions, extreme poses, or low-resolution conditions.

\section{Conclusion and Future Work}
This paper presents HadaSmileNet, a computationally efficient feature fusion framework that directly integrates transformer-based representations with physiologically-grounded D-Markers for genuine smile recognition. By employing parameter-free Hadamard multiplicative fusion, the approach eliminates the training complexities associated with multi-task learning methods while achieving superior performance. Comprehensive evaluation across four benchmark datasets establishes new state-of-the-art results for deep learning approaches, with significant computational advantages including 26\% parameter reduction and simplified training procedures. The systematic analysis of 15 fusion strategies provides valuable insights for the pattern recognition community, making the framework suitable for practical deployment in multimedia data mining applications requiring efficient affective computing abilities.

Future research directions could explore: \textbf{(a)} investigating alternative parameter-free fusion mechanisms for domain knowledge integration, \textbf{(b)} extending the framework to broader emotion recognition applications, \textbf{(c)} exploring cross-modal fusion with audio and contextual signals, and \textbf{(d)} developing end-to-end approaches that automatically learn physiologically-inspired features without manual extraction.


\begin{thebibliography}{10}
\providecommand{\url}[1]{#1}
\csname url@samestyle\endcsname
\providecommand{\newblock}{\relax}
\providecommand{\bibinfo}[2]{#2}
\providecommand{\BIBentrySTDinterwordspacing}{\spaceskip=0pt\relax}
\providecommand{\BIBentryALTinterwordstretchfactor}{4}
\providecommand{\BIBentryALTinterwordspacing}{\spaceskip=\fontdimen2\font plus
\BIBentryALTinterwordstretchfactor\fontdimen3\font minus \fontdimen4\font\relax}
\providecommand{\BIBforeignlanguage}[2]{{%
\expandafter\ifx\csname l@#1\endcsname\relax
\typeout{** WARNING: IEEEtran.bst: No hyphenation pattern has been}%
\typeout{** loaded for the language `#1'. Using the pattern for}%
\typeout{** the default language instead.}%
\else
\language=\csname l@#1\endcsname
\fi
#2}}
\providecommand{\BIBdecl}{\relax}
\BIBdecl

\bibitem{dong2022intentional}
Z.~Dong, G.~Wang, S.~Lu, L.~Dai, S.~Huang, and Y.~Liu, ``Intentional-deception detection based on facial muscle movements in an interactive social context,'' \emph{Pattern Recognition Letters}, vol. 164, pp. 30--39, 2022.

\bibitem{barrett2016works}
P.~H. Barrett, \emph{The Works of Charles Darwin: Vol 23: The Expression of the Emotions in Man and Animals}.\hskip 1em plus 0.5em minus 0.4em\relax Routledge, 2016.

\bibitem{ekman1990duchenne}
P.~Ekman, R.~J. Davidson, and W.~V. Friesen, ``The duchenne smile: Emotional expression and brain physiology: Ii.'' \emph{Journal of personality and social psychology}, vol.~58, no.~2, p. 342, 1990.

\bibitem{wegrzyn2017mapping}
M.~Wegrzyn, M.~Vogt, B.~Kireclioglu, J.~Schneider, and J.~Kissler, ``Mapping the emotional face. how individual face parts contribute to successful emotion recognition,'' \emph{PloS one}, vol.~12, no.~5, p. e0177239, 2017.

\bibitem{sarma2021methods}
D.~Sarma and M.~K. Bhuyan, ``Methods, databases and recent advancement of vision-based hand gesture recognition for hci systems: A review,'' \emph{SN Computer Science}, vol.~2, no.~6, p. 436, 2021.

\bibitem{BRUCE201595}
N.~D. Bruce, C.~Wloka, N.~Frosst, S.~Rahman, and J.~K. Tsotsos, ``On computational modeling of visual saliency: Examining what’s right, and what’s left,'' \emph{Vision Research}, vol. 116, pp. 95--112, 2015, computational Models of Visual Attention.

\bibitem{oh2016let}
S.~Y. Oh, J.~Bailenson, N.~Kr{\"a}mer, and B.~Li, ``Let the avatar brighten your smile: Effects of enhancing facial expressions in virtual environments,'' \emph{PloS one}, vol.~11, no.~9, p. e0161794, 2016.

\bibitem{lander2020recognizing}
K.~Lander and N.~L. Butcher, ``Recognizing genuine from posed facial expressions: exploring the role of dynamic information and face familiarity,'' \emph{Frontiers in Psychology}, vol.~11, p. 1378, 2020.

\bibitem{li2020deep}
S.~Li and W.~Deng, ``Deep facial expression recognition: A survey,'' \emph{IEEE transactions on affective computing}, vol.~13, no.~3, pp. 1195--1215, 2020.

\bibitem{kawulok2021dynamics}
M.~Kawulok, J.~Nalepa, J.~Kawulok, and B.~Smolka, ``Dynamics of facial actions for assessing smile genuineness,'' \emph{Plos one}, vol.~16, no.~1, p. e0244647, 2021.

\bibitem{hassen2021new}
O.~A. Hassen, N.~A. Abu, Z.~Z. Abidin, and S.~M. Darwish, ``A new descriptor for smile classification based on cascade classifier in unconstrained scenarios,'' \emph{Symmetry}, vol.~13, no.~5, p. 805, 2021.

\bibitem{ratnawati2019features}
D.~E. Ratnawati, S.~Anam \emph{et~al.}, ``Features selection for classification of smiles codes based on their function,'' in \emph{ISRITI}.\hskip 1em plus 0.5em minus 0.4em\relax IEEE, 2019, pp. 103--108.

\bibitem{cohn2004timing}
J.~F. Cohn and K.~L. Schmidt, ``The timing of facial motion in posed and spontaneous smiles,'' \emph{International Journal of Wavelets, Multiresolution and Information Processing}, vol.~2, no.~02, pp. 121--132, 2004.

\bibitem{dibeklioglu2010eyes}
H.~Dibeklioglu, R.~Valenti, A.~A. Salah, and T.~Gevers, ``Eyes do not lie: Spontaneous versus posed smiles,'' in \emph{ACM Multimedia}, 2010, pp. 703--706.

\bibitem{pfister2011differentiating}
T.~Pfister, X.~Li, G.~Zhao, and M.~Pietik{\"a}inen, ``Differentiating spontaneous from posed facial expressions within a generic facial expression recognition framework,'' in \emph{ICCV Workshops}.\hskip 1em plus 0.5em minus 0.4em\relax IEEE, 2011, pp. 868--875.

\bibitem{wu2014spontaneous}
P.~Wu, H.~Liu, and X.~Zhang, ``Spontaneous versus posed smile recognition using discriminative local spatial-temporal descriptors,'' in \emph{ICASSP}.\hskip 1em plus 0.5em minus 0.4em\relax IEEE, 2014, pp. 1240--1244.

\bibitem{dibekliouglu2015recognition}
H.~Dibeklio{\u{g}}lu, A.~A. Salah, and T.~Gevers, ``Recognition of genuine smiles,'' \emph{IEEE Transactions on Multimedia}, vol.~17, no.~3, pp. 279--294, 2015.

\bibitem{wu2017spontaneous}
P.-p. Wu, H.~Liu, X.-w. Zhang, and Y.~Gao, ``Spontaneous versus posed smile recognition via region-specific texture descriptor and geometric facial dynamics,'' \emph{Frontiers of Information Technology \& Electronic Engineering}, vol.~18, no.~7, pp. 955--967, 2017.

\bibitem{mandal2017spontaneous}
B.~Mandal and N.~Ouarti, ``Spontaneous versus posed smiles—can we tell the difference?'' in \emph{CVIP 2016, Volume 2}.\hskip 1em plus 0.5em minus 0.4em\relax Springer, 2017, pp. 261--271.

\bibitem{ekman1978facial}
P.~Ekman and W.~V. Friesen, ``Facial action coding system,'' \emph{Environmental Psychology \& Nonverbal Behavior}, 1978.

\bibitem{ekman1993facial}
P.~Ekman, ``Facial expression and emotion.'' \emph{American psychologist}, vol.~48, no.~4, p. 384, 1993.

\bibitem{mandal2017distinguishing}
B.~Mandal, D.~Lee, and N.~Ouarti, ``Distinguishing posed and spontaneous smiles by facial dynamics,'' in \emph{ACCV}.\hskip 1em plus 0.5em minus 0.4em\relax Springer, 2017, pp. 552--566.

\bibitem{yang2020realsmilenet}
Y.~Yang, M.~Z. Hossain, T.~Gedeon, and S.~Rahman, ``Realsmilenet: A deep end-to-end network for spontaneous and posed smile recognition,'' in \emph{Computer Vision -- ACCV 2020}, H.~Ishikawa, C.-L. Liu, T.~Pajdla, and J.~Shi, Eds.\hskip 1em plus 0.5em minus 0.4em\relax Cham: Springer International Publishing, 2021, pp. 21--37.

\bibitem{faroque2022less}
M.~T. Faroque, Y.~Yang, M.~Z. Hossain, S.~M. Naim, N.~Mohammed, and S.~Rahman, ``Less is more: Facial landmarks can recognize a spontaneous smile,'' \emph{arXiv preprint arXiv:2210.04240}, 2022.

\bibitem{fan2022pstnet}
H.~Fan, X.~Yu, Y.~Ding, Y.~Yang, and M.~Kankanhalli, ``Pstnet: Point spatio-temporal convolution on point cloud sequences,'' \emph{arXiv preprint arXiv:2205.13713}, 2022.

\bibitem{fan2021point}
H.~Fan, Y.~Yang, and M.~Kankanhalli, ``Point 4d transformer networks for spatio-temporal modeling in point cloud videos,'' in \emph{CVPR}, 2021, pp. 14\,204--14\,213.

\bibitem{dosovitskiy2020image}
A.~Dosovitskiy, L.~Beyer, A.~Kolesnikov, D.~Weissenborn, X.~Zhai, T.~Unterthiner, M.~Dehghani, M.~Minderer, G.~Heigold, S.~Gelly \emph{et~al.}, ``An image is worth 16x16 words: Transformers for image recognition at scale,'' \emph{arXiv preprint arXiv:2010.11929}, 2020.

\bibitem{HASAN2024148}
M.~J. Hasan, K.~Rafat, F.~Rahman, N.~Mohammed, and S.~Rahman, ``Deepmarkernet: Leveraging supervision from the duchenne marker for spontaneous smile recognition,'' \emph{Pattern Recognition Letters}, vol. 186, pp. 148--155, 2024.

\bibitem{10965485}
G.~G. Chrysos, Y.~Wu, R.~Pascanu, P.~Torr, and V.~Cevher, ``Hadamard product in deep learning: Introduction, advances and challenges,'' \emph{IEEE Transactions on Pattern Analysis and Machine Intelligence}, pp. 1--20, 2025.

\bibitem{grishchenko2020attention}
I.~Grishchenko, A.~Ablavatski, Y.~Kartynnik, K.~Raveendran, and M.~Grundmann, ``Attention mesh: High-fidelity face mesh prediction in real-time,'' \emph{arXiv preprint arXiv:2006.10962}, 2020.

\bibitem{schmidt2003signal}
K.~L. Schmidt, J.~F. Cohn, and Y.~Tian, ``Signal characteristics of spontaneous facial expressions: Automatic movement in solitary and social smiles,'' \emph{Biological psychology}, vol.~65, no.~1, pp. 49--66, 2003.

\bibitem{xiang2021walk}
T.~Xiang, C.~Zhang, Y.~Song, J.~Yu, and W.~Cai, ``Walk in the cloud: Learning curves for point clouds shape analysis,'' in \emph{Proceedings of the IEEE/CVF international conference on computer vision}, 2021, pp. 915--924.

\end{thebibliography}
\end{document}